\documentclass[final]{cvpr}

\usepackage{times}
\usepackage{epsfig}
\usepackage{graphicx}
\usepackage{amsmath}
\usepackage{amssymb}
\usepackage{multirow}
\usepackage{tabularx}
\graphicspath{{graphics/}}
\usepackage[pagebackref=true,breaklinks=true,colorlinks,bookmarks=false]{hyperref}

\def\cvprPaperID{2404} % *** Enter the CVPR Paper ID here
\def\confYear{CVPR 2021}

\begin{document}

%%%%%%%%% TITLE
\title{Embracing Uncertainty: Decoupling and De-bias for Robust Temporal Grounding}

\author{Hao Zhou\textsuperscript{1}, Chongyang Zhang\textsuperscript{1,2} \thanks{This is the corresponding author.}, Yan Luo\textsuperscript{1}, Yanjun Chen\textsuperscript{1}, Chuanping Hu\textsuperscript{1,3}\\
\textsuperscript{1}School of Electronic Information and Electrical Engineering, Shanghai Jiao Tong University\\
\textsuperscript{2}MoE Key Lab of Artificial Intelligence, AI Institute, Shanghai Jiao Tong University, Shanghai, China\\
\textsuperscript{3}Zhengzhou University, Zhengzhou, China\\
{\tt\small \{zhouhao\_0039,sunny\_zhang,luoyan\_bb,erinchen\}@sjtu.edu.cn, cphu@vip.sina.com}
% For a paper whose authors are all at the same institution,
% omit the following lines up until the closing ``}''.
% Additional authors and addresses can be added with ``\and'',
% just like the second author.
% To save space, use either the email address or home page, not both
% \and
% Second Author\\
% Institution2\\
% First line of institution2 address\\
% {\tt\small secondauthor@i2.org}
}

\maketitle
% \thispagestyle{empty}

%%%%%%%%% ABSTRACT
\begin{abstract}
Temporal grounding aims to localize temporal boundaries within untrimmed videos by language queries, but it faces the challenge of two types of inevitable human uncertainties: query uncertainty and label uncertainty. The two uncertainties stem from human subjectivity, leading to limited generalization ability of temporal grounding. In this work, we propose a novel DeNet ($\textbf{De}$coupling and $\textbf{De}$-bias) to embrace human uncertainty: Decoupling — We explicitly disentangle each query into a relation feature and a modified feature. The relation feature, which is mainly based on skeleton-like words (including nouns and verbs), aims to extract basic and consistent information in the presence of query uncertainty. Meanwhile, modified feature assigned with style-like words (including adjectives, adverbs, etc) represents the subjective information, and thus brings personalized predictions; De-bias —  We propose a de-bias mechanism to generate diverse predictions, aim to alleviate the bias caused by single-style annotations in the presence of label uncertainty. Moreover, we put forward new multi-label metrics to diversify the performance evaluation. Extensive experiments show that our approach is more effective and robust than state-of-the-arts on Charades-STA and ActivityNet Captions datasets.
\end{abstract}

%%%%%%%%% BODY TEXT
\section{Introduction}

As the increasing demand for video understanding, many related works have drawn increasing attention, \eg action recognition~\cite{2014Two,2016Temporal,liu2018global} and temporal action detection~\cite{Zhao_2017_ICCV,lin2017single}. These tasks rely on trimmed videos or predefined action categories, yet most videos are untrimmed and associated with open-world language descriptions in real scenarios. Temporal grounding task aims to localize corresponding temporal boundaries in an untrimmed video by a language query. Thus, models need to understand both fine-grained video content and complex language queries. Recently, this task has also shown its potential in a wide range of applications, \eg video captioning~\cite{pan2016jointly,Yu_2016_CVPR,duan2018weakly}, video object segmentation~\cite{Gavrilyuk_2018_CVPR,khoreva2018video} and video question answering~\cite{lei2018tvqa,kim2019gaining,wang2019holistic}.

\begin{figure}[t]
\begin{center}
   \includegraphics[width=1\linewidth]{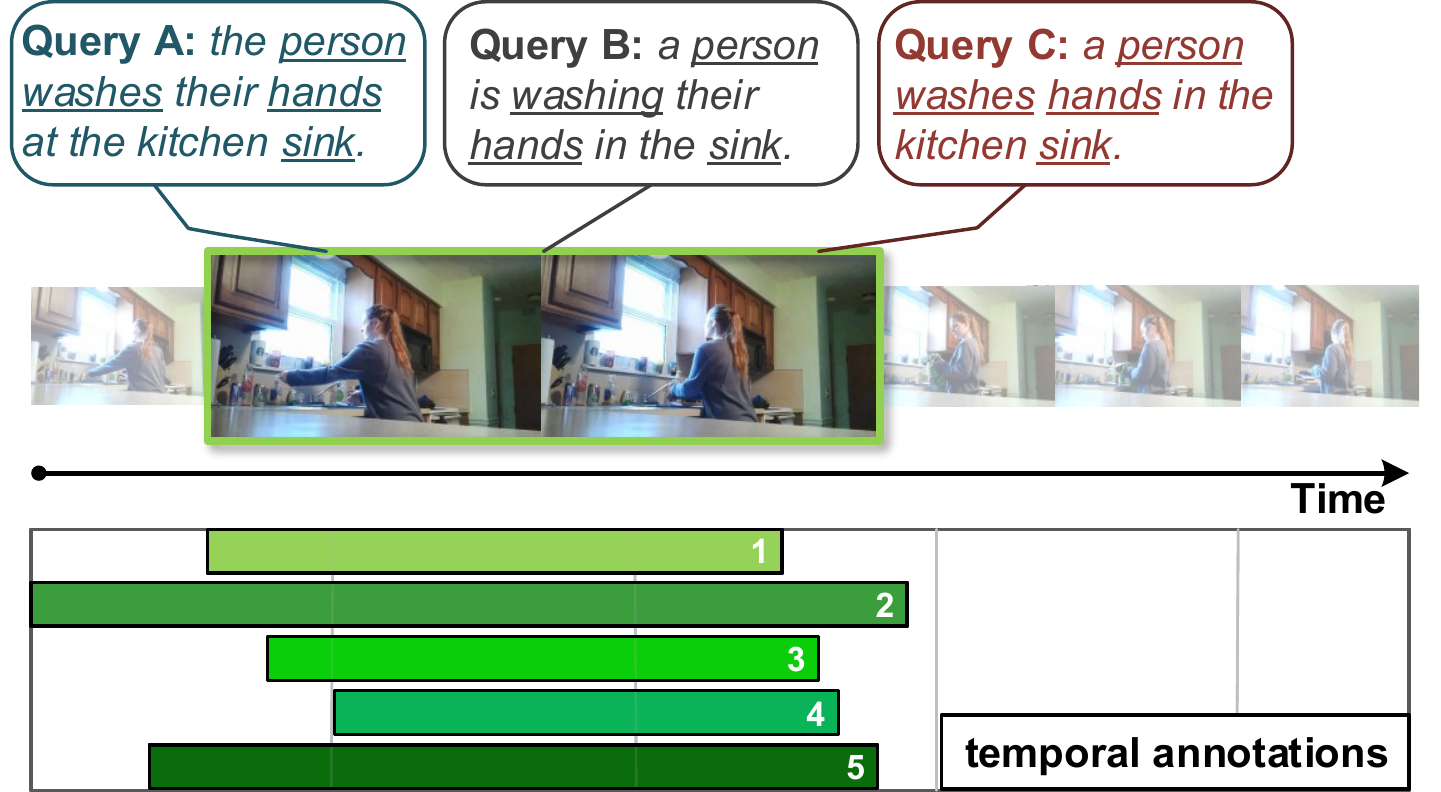}
\end{center}
   \caption{Example of temporal grounding task with two types of uncertainties. Query uncertainty: For one same event, there are different language expressions. Label uncertainty: Given one same query and video, different annotators may provide a variety of temporal boundaries.}
\label{fig:1}
\end{figure}

We observe there lies inherent uncertainty in temporal grounding task and classify it into two types: 1) One is query uncertainty stemming from different expressions for one same event. As shown in Figure~\ref{fig:1}, three queries are attached to the same moment. Previous approaches usually leverage LSTM-based~\cite{zeng2020dense,2DTAN_2020_AAAI} networks to encode entire language as a deterministic vector. However, the variety of expressions makes it challenging to extract discriminative semantic features, sometimes leading to quite different predictions for the same event. 
2) The other is label uncertainty representing subjective boundaries for one same event. As shown in Figure~\ref{fig:1}, for the same query A and video, temporal boundaries annotated by different people exist disagreement. Due to the expensive cost of multiple-labeling, most of previous models~\cite{mun2020local,rodriguez2020proposal} are optimized using single-style annotations (which means each sample is labeled by one annotator), whereas the inherent uncertainty of event localization~\cite{sigurdsson2017actions} is ignored. As a result, models may learn single-style prediction bias from training datasets, leading to limited generalization performances. 

Considering the fact that uncertainty can cover a broad range of human perspectives, it should be embraced to promote robust temporal grounding. Furthermore, we argue single-annotation, single prediction is not reasonable in the presence of uncertainty, and diversity of predictions is an effective way to alleviate the bias caused by single-style annotations. Therefore, the key challenge is how to obtain diverse predictions. 
Inspired by linguistic knowledge, we find consistent discriminative information lies in a skeleton-like relation phrase (including {\em nouns} and {\em verbs}), and query uncertainty mainly exists in a style-like modified phrase (including {\em adjectives}, {\em adverbs}, etc). On one hand, the relation phrase is beneficial to robust temporal grounding. On the other hand, the modified phrase may be largely associated with human preferences and brings personalized differences. Based on this intuition, our main idea is to leverage various expressions stemming from query uncertainty to obtain a diverse yet plausible prediction set that fits label uncertainty.

In this paper, we propose one novel DeNet ($\textbf{De}$coupling and $\textbf{De}$-bias) to embrace the two types of uncertainties in the temporal grounding task. First of all, a decoupling method is introduced to disentangle each query into a relation feature and a modified feature using Parts-of-Speech (PoS). While discriminative and consistent information is obtained from the relation feature, personalized information can be also reserved in the modified feature. Then, a de-bias mechanism is proposed to generate diverse predictions, which includes sampling operation, multiple choice learning (MCL)~\cite{guzman2012multiple}, clustering, etc. Specifically, we encode the modified feature as a Gaussian distribution and adopt a sampling operation in the latent space to obtain multiple query representations. To tackle the dilemma between multiple predictions and single-style annotations, we introduce a min-loss from MCL to optimize DeNet to generate diverse predictions. In the inference stage, multiple predictions are clustered into one diverse yet plausible prediction set. Moreover, we devise multi-label metrics to meet for multiple testing annotations situations. Finally, DeNet is evaluated on two popular datasets Charades-STA~\cite{gao2017tall} and ActivityNet Captions~\cite{caba2015activitynet,krishna2017dense} in terms of standard metrics and new multi-label metrics.
To sum up, the main contributions of our work are as follows:

(1) We first attempt to embrace two types of human uncertainties: query uncertainty and label uncertainty, in one unified network DeNet to model robust temporal grounding.

(2) We develop a decoupling module in the language encoding, and one de-bias mechanism in the temporal regression. With the two designs, diverse yet plausible predictions can be obtained to fit human diversity in real scenarios.

(3) We devise new multi-label metrics to meet multiple annotations and verify the effectiveness and robustness of DeNet on both Charades-STA and ActivityNet Captions.

%-------------------------------------------------------------------------
\section{Related Work}
\noindent
{\bf Temporal grounding.} 
As a challenging task in video understanding, temporal grounding needs to capture semantic information in both videos and language queries.

In the video encoding component, most previous approaches~\cite{gao2017tall,liu2018attentive,anne2017localizing,wang2019language,2DTAN_2020_AAAI} follow a proposal-based framework, where untrimmed videos are clipped into multi-scale segments as proposal candidates. Gao \etal~\cite{gao2017tall} and Liu \etal~\cite{liu2018attentive} adopt a sliding window to combine each central-clip feature and its context-clip features as one proposal candidate. Hendricks \etal~\cite{anne2017localizing} and Wang \etal~\cite{wang2019language} concatenate local features and global feature to better cover contexts. To further explore dependencies across multiple candidates, Zhang \etal~\cite{2DTAN_2020_AAAI} generate multi-scale segments and construct a 2D temporal adjacent map. However, too many proposals will burden models during the training process. Recently, some approaches~\cite{mun2020local,zeng2020dense,rodriguez2020proposal,he2019read} adopt a proposal-free framework. For example, Zeng \etal~\cite{zeng2020dense} extract sequential clip-level features, then directly predict temporal boundaries in a subsequent network. In this paper, the proposed DeNet follows the proposal-free framework to reduce the training computation cost.

Language encoding also plays an important role in the temporal grounding task. Most approaches employ LSTM-based layers~\cite{zeng2020dense,2DTAN_2020_AAAI,mun2020local,xu2019multilevel} or GRU-based layers~\cite{rodriguez2020proposal,yuan2019semantic} to encode entire language queries. Recently, some approaches~\cite{zhang2019cross,zhang2019exploiting,liu2018temporal} leverage syntactic dependency parser to capture underlying semantic structures. Besides, Mun \etal~\cite{mun2020local} and Yuan \etal~\cite{yuan2019find} attempt to capture discriminative features from queries using an attention mechanism. These methods aim to obtain more subtle query representations, yet we follow a different motivation. On the one hand, we hope to obtain discriminative information from various expressions to achieve robust predictions. On the other hand, we attempt to reserve personalized differences to achieve diversified predictions. Thus, we adopt an explicit decoupling method to disentangle each query into the relation feature and the modified feature. 

\begin{figure*}
\begin{center}
\includegraphics[width=0.95\linewidth]{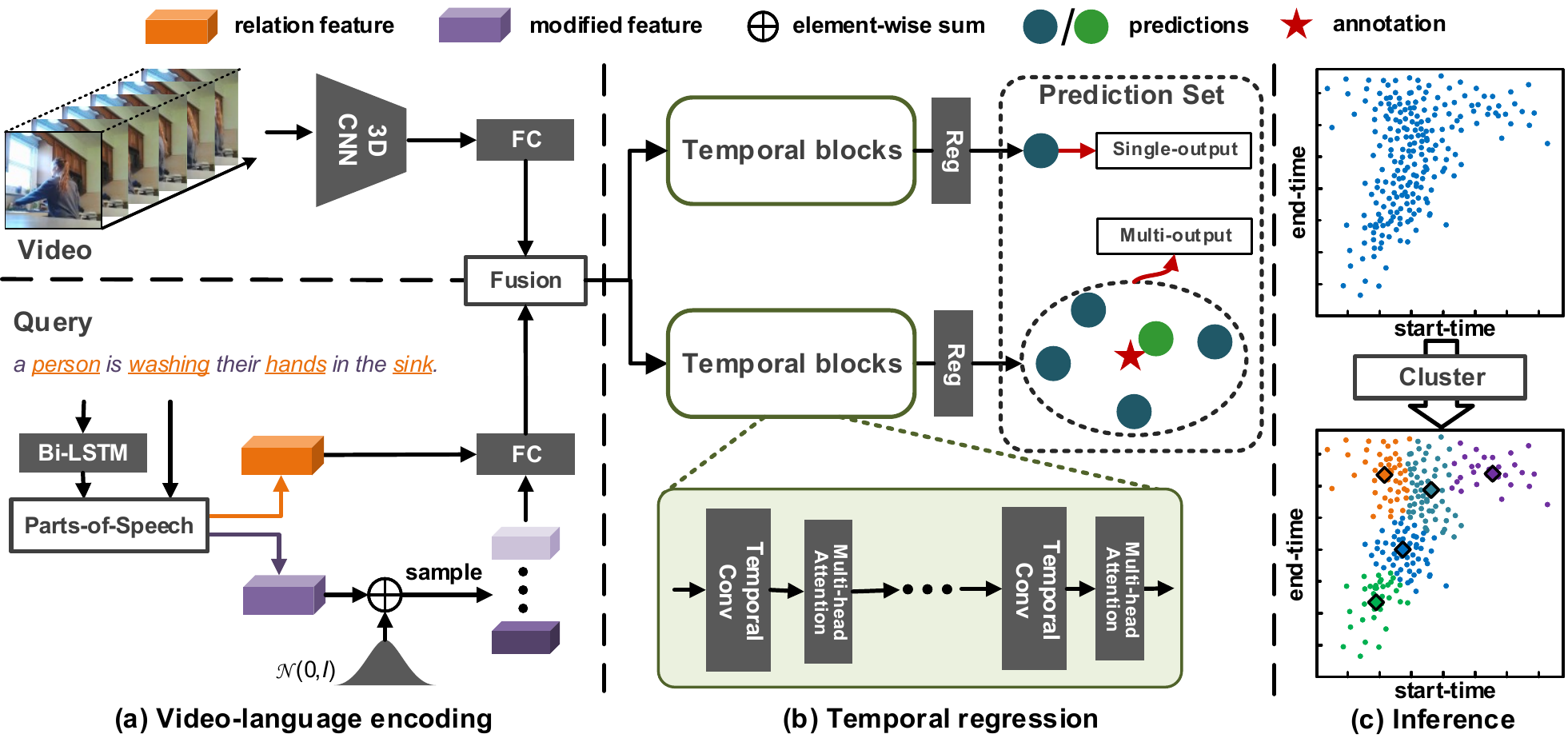}
\end{center}
   \caption{An overview of our proposed model for the temporal grounding task. (a) In the video-language encoding component, we use a pretrained 3D CNN to extract the sequential video feature and disentangle the query into relation feature and modified feature by Parts-of-Speech. Then, a sampling operation is applied in the latent space to generate multiple query representations. (b) In the temporal regression component, two independent branches are set to generate multiple predictions. (c) In the inference stage, we adopt a clustering method to obtain a fixed-size prediction set.}
\label{fig:2}
\end{figure*}

% \noindent
% {\bf Learning with uncertainty.} 
% Human uncertainty arises from the subjectivity of human annotators, which has been researched in various areas. In the saliency detection task, Zhang \etal~\cite{zhang2020uc} construct a conditional variational auto-encoder to model the uncertainty. Through sampling in the latent space, they provide multiple saliency maps for each input image. In the field of Person Re-identification, Yu \etal~\cite{yu2019robust} propose a distribution network to represent each person image as a Gaussian distribution. Kohl \etal~\cite{kohl2018probabilistic} produce a collection of diverse segmentation to handle uncertainty in the image segmentation task. 

% However, few works have explored human uncertainty in the temporal grounding task. Recently, Rodriguez \etal~\cite{rodriguez2020proposal} attempt to add a Gaussian kernel to annotations as soft labels, yet performances are still limited.

\noindent
{\bf Multiple choice learning.} 
In contrast to single-output learning, multiple choice learning (MCL)~\cite{guzman2012multiple} is proposed to produce multiple outputs based on one min-loss. Given a training sample, MCL takes account of all hypotheses and only updates networks according to the best hypothesis. One accurate and diverse prediction set can be obtained in this way. Inspired by MCL, we consider diversity is an effective way to model human uncertainty, and introduce the min-loss into temporal grounding to predict all possible temporal boundaries in the absence of multiple annotations. However, note that our proposed method is significantly different from traditional MCLs. Firstly, MCL focuses on ensemble learning, whereas we focus on temporal grounding. Then, most MCL approaches~\cite{lee2016stochastic,lee2017confident,Tian2019VersatileMC} produce the multi-output $\{f_i( x)\}_{i=1}^{N}$ based on multiple "base classifiers", whereas our method generates the multi-output $\{f(\hat x_i)\}_{i=1}^{N}$ via multiple features.

%-------------------------------------------------------------------------
\section{Proposed Method}
\subsection{Method overview}
Given an untrimmed video $\mathrm{\bf V}$ and an open-world language description $\mathrm{\bf Q}$ as a query, temporal grounding aims to localize the start-end boundary $\mathrm{\bf b}_{se}$ within $\mathrm{\bf V}$. Specifically, the untrimmed video is represented as $\mathrm{\bf V} = \{\mathrm{\bf v}_i\}_{i=1}^T$, where $\mathrm{\bf v}_i$ denotes the i-th video clip and $T$ is the total number of video clips. The query is represented as $\mathrm{\bf Q} = \{\mathrm{\bf w}_i\}_{i=1}^S$, where $\mathrm{\bf w}_i$ denotes the i-th word and $S$ is the total number of words. In this work, models should output matched temporal times $\{\mathrm{\bf b}_{se}\}^N = \{(t_s, t_e)\}^N$ corresponding to the query $\mathrm{\bf Q}$, where $N$ is the number of predictions.

As illustrated in Figure~\ref{fig:2}, DeNet contains two main components: video-language encoding and temporal regression. In the video-language encoding component, we adopt a decoupling method to disentangle each query into a relation feature and a modified feature using PoS, where the modified feature is encoded as a distribution. Then, the video-language feature is fed into the temporal regression component to predict multiple temporal boundaries. In the training stage, two independent branches are optimized by single-output loss and multi-output loss, respectively. In the inference stage, we cluster the collection of predictions into a fixed-size prediction set and evaluate them in both standard metrics and new multi-label metrics.

\subsection{Video-language encoding}
\noindent
{\bf Video encoding.}  
Firstly, an untrimmed video is represented as a collection of clips $\mathrm{\bf V} = \{\mathrm{\bf v}_i\}_{i=1}^T$, where each clip covers $C$ frames ($C = 16$ in this work). Analogous to \cite{2DTAN_2020_AAAI}, we use a pretrained 3D CNN model to extract clip-level features, then sample fixed $T_m$ clips from $T$ clips so as to obtain a fixed-length video feature $\mathrm{\bf \widetilde{V}} \in \mathbb{R}^{d_v\times {T_m}}$, where $d_v$ is the dimension of the video feature. Furthermore, a zero-padding operation is applied if there are less than $T_m$ clips in an untrimmed video. Finally, two extra Fully Connected layers are implemented to obtain a final video embedding $\mathrm{\bf F}^V \in \mathbb{R}^{d_v\times {T_m}} $ as:

\begin{equation}
\mathrm{\bf F}^V  = \mathrm{\bf W}_2\mathrm{ ReLU}(\mathrm{\bf W}_1\mathrm{\bf \widetilde{V}}),
\label{equ:1}
\end{equation}
where $\mathrm{\bf W}_1$, $\mathrm{\bf W}_2 \in \mathbb{R}^{d_v\times {d_v}}$ are learnable parameters, the superscript $V$ indicates the video modality.

\noindent
{\bf Language encoding.} 
For a language query $\mathrm{\bf Q} = \{\mathrm{\bf w}_i\}_{i=1}^S$ with $S$ words, we take advantage of Glove~\cite{pennington2014glove} to map each word to a 300-dimensional vector, then set two Bi-LSTM layers to get word-level features $\{\mathrm{\bf h}_i\}_{i=1}^S \in \mathbb{R}^{d_l\times {S}}$, where $d_l$ is the feature dimension of each word. In our observation, query uncertainty mainly lies in the modified phrase and discriminative information are in the relation phrase. For example, "{\em a person is washing their hands in the sink}" can be broken down into relation phrase [{\em person, washing, hands, sink}] and modified phrase [{\em a, is, their, in, the}]. 
Here, the spaCy toolbox\footnote {https://spacy.io/} is used to generate PoS tags that denote word types, like {\em verbs}, {\em adjectives}. Then, we average word-level features associated with the relation phrase to get a relation feature $\mathrm{\bf f}^L_r$. Similarly, the remaining word-level features are selected and averaged as a modified feature $\mathrm{\bf f}^L_m$.

Then, we concatenate the two types of features and set a Fully Connected layer to obtain a final query embedding as:
\begin{equation}
\mathrm{\bf f}^L  = \mathrm{\bf W}_3[\mathrm{\bf f}^L_r,\mathrm{\bf f}^L_m] + \mathrm{\bf b}_3,
\label{equ:2}
\end{equation}
where $\mathrm{\bf W}_3 \in \mathbb{R}^{d_l \times 2d_l}$, $\mathrm{\bf b}_3 \in \mathbb{R}^{d_l}$ are the learnable parameters, $[\cdot, \cdot]$ denotes concatenation and the superscript $L$ indicates the language modality. 
Considering the fact that most variances stem from the modified phrase, we encode corresponding modified feature as a distribution instead of a deterministic vector. Here, we adopt the Gaussian distribution $\mathcal N (\mathrm{\bf u}, {\bf \sigma}^2)$ as in many existing works~\cite{yoo2019data}. From a probabilistic perspective, it means that the feature is regarded as a random variable to model uncertainty~\cite{yu2019robust}. $\mathrm{\bf f}^L_m$ is set as the distribution center $\bf u$ and a collection of modified features are sampled from the Gaussian distribution $\mathcal N (\mathrm{\bf f}^L_m, {\bf \sigma}^2)$. A reparameterisation trick is used to obtain the modified feature $\mathrm{\bf \hat f}^L_m = \mathrm{\bf f}^L_m + \epsilon, \epsilon \sim \mathcal N (\bf 0, {\bf \sigma}^2)$. Finally, a variant query embedding is formulated as:
\begin{equation}
\mathrm{\bf \hat f}^L  = \mathrm{\bf W}_4[\mathrm{\bf f}^L_r, \mathrm{\bf \hat f}^L_m] + \mathrm{\bf b}_4.
\label{equ:3}
\end{equation}

From another perspective, the distribution representation is equivalent to adding small perturbations in the modified feature. We provide two rationales illustrating its advantages. On the one hand, models will further focus on the relation feature and pay less attention to the modified feature. Thereby, the model is more robust. On the other hand, the sampling process can be viewed as query augmentation. Based on multiple query features, models can generate multiple personalized predictions.

\noindent
{\bf Multimodal fusion.} 
When both videos and language embeddings are obtained, we need to model the interaction of them. First of all, $\mathrm{\bf f}^L $ and $\mathrm{\bf \hat f}^L $ are replicated for $T_m$ times to get sequential embeddings $\mathrm{\bf F}^L$, $\mathrm{\bf \hat F}^L \in \mathbb{R}^{d_l\times {T_m}}$, respectively. Then, multimodal features $\mathrm{\bf F}^M, \mathrm{\bf \hat F}^M \in \mathbb{R}^{d_m\times {T_m}}$ are produced by fusing video embedding and query embedding:
\begin{equation}
\mathrm{\bf F}^M  = || \mathrm{\bf F}^V \circ \mathrm{\bf F}^L||_F,
\label{equ:4}
\end{equation}
\begin{equation}
\mathrm{\bf  \hat F}^M  = ||\mathrm{\bf F}^V \circ \mathrm{\bf \hat F}^L||_F,
\label{equ:5}
\end{equation}
where $\circ$ denotes the Hadamard product and $||\cdot ||_F$ is the Frobenius normalization ($\ell_2$-norm). Note that $d_m$, $d_v$ and $d_l$ are consistent for dimension matching.

\subsection{Temporal regression}
When we obtain a collection of multimodal features, a temporal regression network is constructed to predict matched temporal boundaries. It is composed of two independent branches, where each branch contains a stack of temporal blocks and a regression layer. The single-output branch associated with $\mathrm{\bf F}^M $ produces a top-1 prediction. The multi-output branch associated multiple $\mathrm{\bf \hat F}^M$ produces multiple predictions covering possible annotations.

Each temporal block contains a Temporal Convolutional layer and a Multi-head Attention layer~\cite{vaswani2017attention}. The Temporal Convolutional layer aims to capture temporal dependencies in the neighbor clips and the Multi-head Attention layer is to capture long-range temporal dependencies. For the n-th temporal block, its output $\mathrm{\bf F}^{(n)} \in \mathbb{R}^{d_m\times {T_m}}$ can be formulated as:
\begin{equation}
\mathrm{\bf \widetilde{F}}^{(n)}  = \mathrm{\bf F}^{(n-1)} + \mathrm{Conv}(\mathrm{\bf F}^{(n-1)}),
\label{equ:6}
\end{equation}
\begin{equation}
\mathrm{\bf F}^{(n)}  = \mathrm{\bf \widetilde{F}}^{(n)} +\mathrm{MultiheadAttention}(\mathrm{\bf \widetilde{F}}^{(n)}),
\label{equ:7}
\end{equation}
where $\mathrm{\bf F}^{(n-1)}$ is the output of previous temporal block. $\mathrm{Conv}(\cdot)$ represents a mapping function in the Temporal Convolutional layer that contains two 1D convolutional layers with batch normalization. 

Following a stack of temporal blocks, an attention-guided regression layer is employed to output the start-end prediction $\mathrm{\bf b}_{se}$. An auxiliary head is implemented here to predict the center-width $\mathrm{\bf b}_{cw}$ to assist temporal grounding. Thus, the regression layer is formulated as:
\begin{equation}
\mathrm{\bf a}  = \mathrm{softmax}(\mathrm{\mathrm{\bf W}_{6}}\mathrm{Tanh}(\mathrm{\bf W}_{5}\mathrm{\bf F} )),
\label{equ:8}
\end{equation}
\begin{equation}
\mathrm{\bf b}_{cw} = (t_c, t_w)  = \mathrm{\bf Reg}_{cw}(\sum_{i=1}^{T_m} a_i\mathrm{\bf F}_i),
\label{equ:9}
\end{equation}
\begin{equation}
\mathrm{\bf b}_{se} = (t_s, t_e)  = \mathrm{\bf Reg}_{se}(\sum_{i=1}^{T_m} a_i\mathrm{\bf F}_i),
\label{equ:10}
\end{equation}
where $\mathrm{\bf a} \in \mathbb{R}^{{T_m}}$ is an attention coefficient and $\mathrm{\bf Reg}_{cw}$, $\mathrm{\bf Reg}_{se}$ are two independent Fully Connected layers. Note that all of predictions are normalized to [0,1].

\subsection{Optimization and inference}
\noindent
{\bf Optimization.} 
According to the definition of equation~\ref{equ:6}-\ref{equ:10}, we feed $\mathrm{\bf F}^M$ and the collection of $\mathrm{\bf \hat F}^M$ into the two branches of temporal regression network and obtain a single prediction $(\mathrm{\bf b}_{se},\mathrm{\bf b}_{cw},\mathrm{\bf a})$ and multiple predictions $\{(\mathrm{\bf \hat b}_{se},\mathrm{\bf\hat b}_{cw},\mathrm{\bf \hat a})\}^K$, respectively.

The single-output branch is optimized with two kinds of loss functions, and one is a regression loss as follows:
\begin{equation}
\mathcal{ L}_{reg}(\mathrm{\bf b}_{se},\mathrm{\bf b}_{cw})= L_1(\mathrm{\bf b}_{se} - \mathrm{\bf y}_{se}) + L_1(\mathrm{\bf b}_{cw} - \mathrm{\bf y}_{cw}),
\label{equ:11}
\end{equation}
where $L_1$ denotes L1 distances, and $\mathrm{\bf y}_{se}$, $\mathrm{\bf y}_{cw} \in [0,1]$ denote the start-end and center-width groundtruth, respectively. The other one is an attention loss~\cite{yuan2019find} that forces the model to focus on clips within groundtruth interval:

\begin{equation}
\mathcal{ L}_{att}(\mathrm{\bf a})= -\frac{\sum_{i=1}^{T_m} m_i \mathrm{log} a_i}{\sum_{i=1}^{T_m} m_i},
\label{equ:12}
\end{equation}
where $m_i = 1 $ if the i-th clip is within the groundtruth interval and otherwise $m_i = 0 $.

For the multi-output branch, it's not reasonable to regress all of $\{(\mathrm{\bf \hat b}_{se},\mathrm{\bf\hat b}_{cw},\mathrm{\bf \hat a})\}^K$ with one single annotation. To tackle the dilemma between multiple predictions and single-style annotations, we introduce a min-loss from MCL~\cite{guzman2012multiple} to learn diverse predictions without extra annotations. It only computes a loss between the closest prediction to the existing annotations. Finally, all of the loss functions are jointly considered as follows:
\begin{equation}
\begin{aligned}
\mathcal{L}_{all} &= \mathcal{L}_{single} + \lambda \mathcal{L}_{multi}\\
& = \mathcal{ L}_{reg}(\mathrm{\bf b}_{se},\mathrm{\bf b}_{cw}) + \mathcal{ L}_{att}(\mathrm{\bf a}) \\
&\quad + \lambda \mathop{\mathrm{min}}_{i \in [K]} [\mathcal{ L}_{reg}(\mathrm{\bf \hat b}_{se,i},\mathrm{\bf \hat b}_{cw,i}) + \mathcal{ L}_{att}(\mathrm{\bf \hat a}_i)],
\end{aligned}
\label{equ:13}
\end{equation}
where $\lambda$ is a trade-off parameter between two regression branches, and $[K]$ denotes the set $\{1,..., K\}$.

\begin{figure}[t]
\begin{center}
   \includegraphics[width=1\linewidth]{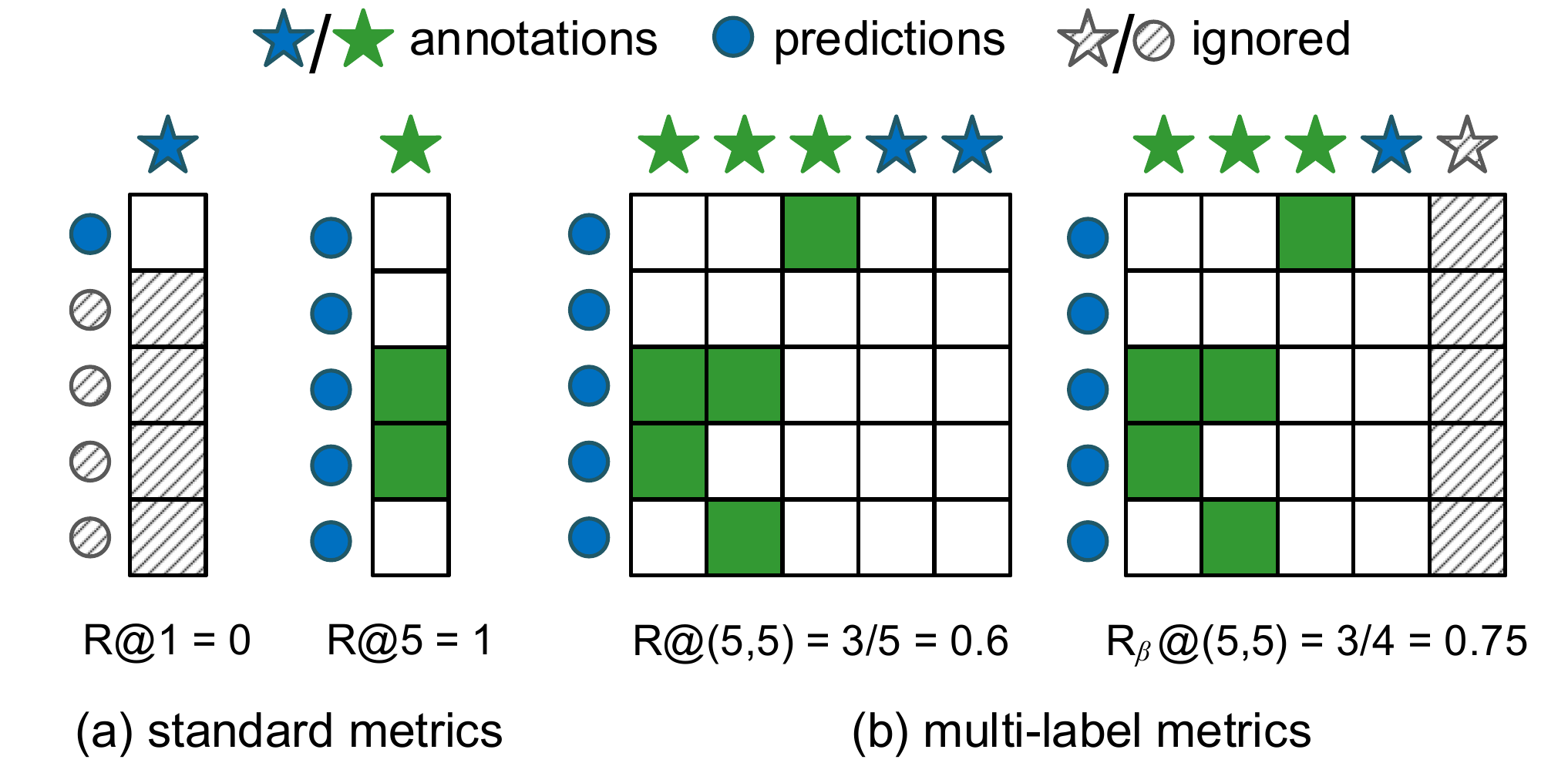}
\end{center}
   \caption{A example to illustrate differences between proposed multi-label metrics and standard metrics. The green star denotes the corresponding annotation that is matched with at least one prediction (with IoU larger than $\alpha$) and otherwise the corresponding star is blue. The grey star is the low-quality annotation (with average IoU smaller than $\beta$).}
\label{fig:3}
\end{figure}
\noindent
{\bf Inference.} 
We only focus on the single prediction $\mathrm{\bf b}_{se}$ and the collection of predictions $\{\mathrm{\bf \hat b}_{se}\}^K$ in the inference stage, where $K$ depends on the number of query embeddings $\mathrm{\bf \hat F}^M$ sampled in the latent space. Previous approaches adopt NMS to reduce predictions, yet this method faces two issues: 1) Since the collection of predictions is dense, predictions are mistakenly suppressed easily. 2) Confidence scores are necessary to rank predictions. Most approaches build up an extra branch to predict the confidence scores or IoU scores, whereas performances are limited. To address above two issues, we leverage K-Means to cluster $\{\mathrm{\bf \hat b}_{se}\}^K$ into a fixed-size prediction set $\{\mathrm{\bf \hat b}_{se}\}^N$ without NMS, where $N$ is a pre-defined constant. If necessary, we can rank $\{\mathrm{\bf \hat b}_{se}\}^N$ using the distance from single-style prediction $\mathrm{\bf b}_{se}$.

\subsection{New evaluation metrics}
\noindent
The standard evaluation metric is "R@$N$, IoU=$\alpha$". It is defined as the percentage of at least one of the top-$N$ predictions having IoU larger than $\alpha$. This metric only focuses on whether the single groundtruth is localized successfully. Due to the label uncertainty, different people localize various moment boundaries for the same query. That is to say, there are multiple acceptable labels for each query. Thus, we consider the prediction set should be evaluated with multi-labels instead of a single label. 

Recently, Otani \etal~\cite{otani2020uncovering} provide 5 annotations for each testing sample on two public datasets. We propose two multi-label metrics to meet for multi-label situations. The first metric is "R@($N,G$), IoU=$\alpha$" that evaluates performances with $N$ predictions and $G$ annotations for each query. It is defined as the percentage of annotations that match at least one prediction (with IoU larger than $\alpha$) in top-$N$ predictions. This metric is equivalent to the standard metric "R@$N$, IoU=$\alpha$" if $G$ is set as 1. The second metric is "R$_\beta$@$(N,G)$, IoU=$\alpha$", where low-quality annotations (with average IoU among annotations smaller than $\beta$) are ignored. Intuitively, when one annotation has a small average IoU, it tends to be low-quality. Thus, "R$_\beta$@$(N,G)$, IoU=$\alpha$" is equivalent to "R@($N,G$), IoU=$\alpha$" if $\beta$ is set as 0. When there is only one testing sample, Figure~\ref{fig:3} illustrates the results in different metrics. The standard metrics only compute the matched percentage of single annotation (\eg R@1 = 0 and R@5 = 1), our multi-label metrics considers whether multiple annotations are matched (\eg R@(5,5) = 0.6 and R$_\beta$@(5,5) = 0.75). We note that some methods~\cite{anne2017localizing,hendricks2018localizing} consider multiple annotations based on standard metrics that use their aggregator over three out of the four human annotators. 
Similar to our proposed "R$_\beta$@$(N,G)$, IoU=$\alpha$", they ignore part of multi-labels when evaluating. However, instead of discarding one of four labels that has the lowest evaluation score, we evaluate the disagreements among labels and filter out low-quality labels adaptively.
%-------------------------------------------------------------------------
\section{Experiments}
\subsection{Datasets}
\noindent
{\bf Charades-STA.} 
This dataset contains 9,848 videos built on the Charades dataset~\cite{sigurdsson2016hollywood}. Gao \etal~\cite{gao2017tall} provide single temporal annotation for each language query as Charades-STA, where 12,408 samples are split into the training set and 3,720 samples are into the testing set. Recently, Otani \etal~\cite{otani2020uncovering} extend 5 temporal annotations for each query (1,000 queries totally) in the testing set.

\noindent
{\bf ActivityNet Captions.} 
This dataset~\cite{caba2015activitynet} contains 19,209 videos, which was originally proposed by~\cite{krishna2017dense} for dense video captioning task. As the largest dataset in temporal grounding task, it contains 37,417, 17,505, and 17,031 samples for the training set, val\_1 set, and val\_2 set. Due to the lack of of the testing set, we follow a popular split method~\cite{xu2019multilevel} that combines the two validation sets as the testing set. Besides, Otani \etal~\cite{otani2020uncovering} extend 5 temporal annotations for each query (1,288 queries totally) in the validation sets.
\subsection{Implementation details}
In the video encoding, we use pretrained 3D CNN networks to extract clip-level features, where each clip contains 16 consecutive frames. Following previous works, we adopt I3D features~\cite{carreira2018action} for Charades-STA and C3D features~\cite{tran2015learning} for ActivityNet Captions. The max video length $T_m$ is set as 128. In the language encoding, we draw 5 samples $\mathrm{\bf \hat F}^M$ from the latent space and set the standard deviation $\sigma$ as an identical matrix {\em I} during the training procedure. For dimension matching, dimension of video embedding $d_v$, dimension of query embedding $d_l$ and dimension of multimodal feature $d_m$ are all set as 512. In the inference procedure, we set deviation $\sigma$ as 2{\em I} to enlarge the personalized differences from modified feature, and cluster about 200 results into fixed 5 predictions using K-means. The trade-off parameter $\lambda$ in Equation~\ref{equ:13} is set as 0.02. In all experiments, we use Adam~\cite{kingma2014adam} and batch size of 32 for optimization.

\begin{table}
\begin{center}
\setlength{\tabcolsep}{1.5pt}
{
\begin{tabular}{l|c|cccc}
\hline
\multirow{2}{*}{ Method } & \multirow{2}{*}{ Feature } & R@1 & R@1 & R@5 & R@5 \\
                             &         & IoU=0.5 & IoU=0.7 & IoU=0.5 & IoU=0.7 \\
\hline\hline
CTRL~\cite{gao2017tall} & C3D & 23.63 & 8.89 & 58.92 & 29.52\\
SMRL~\cite{wang2019language} & C3D & 24.36 & 11.17 & 61.25 & 32.08\\
MAC~\cite{ge2019mac} & C3D & 30.48 & 12.20 & 64.84 & 35.13\\
MLVI~\cite{xu2019multilevel} & C3D & 35.60 & 15.80 & 79.40 & 45.40\\
CBP~\cite{wang2020temporally} & C3D & 36.80 & 18.87 & 70.94 & 50.19\\
SAP~\cite{chen2019semantic} & VGG & 27.42 & 13.36 & 66.37 & 38.15\\
MAN~\cite{zhang2019man} & VGG & 41.24 & 20.54 & 83.21 & 51.85\\
2D-TAN~\cite{2DTAN_2020_AAAI} & VGG & 42.80 & 23.25 & 80.54 &54.14\\
EXCL~\cite{ghosh2019excl} & I3D & 44.10 & 22.40 & - & -\\
TMLGA~\cite{rodriguez2020proposal} & I3D & 52.02 & 33.74& - & -\\
DRN~\cite{zeng2020dense} & I3D & 53.09 & 31.75 & \underline{89.06} & \underline{60.05}\\
SCDM~\cite{yuan2019semantic} & I3D & 54.44 & 33.43 & 74.43 & 58.08\\
LGI~\cite{mun2020local} & I3D & \underline{59.46} & \underline{35.48} & - & -\\
DeNet(ours) & I3D &\bf 59.70 & \bf 38.52 & \bf 91.24 & \bf 66.83 \\
\hline
\end{tabular}
}
\end{center}
\caption{Comparison with state-of-the-art methods on Charades-STA using standard metrics; bold font indicates best results, underlined second-best.}
\label{tab:1}
\end{table}

\begin{table}
\begin{center}
\setlength{\tabcolsep}{4pt}
{
\begin{tabular}{l|cccc}
\hline
\multirow{2}{*}{ Method } & R@1 & R@1 & R@5 & R@5 \\
                             & IoU=0.3 & IoU=0.5 & IoU=0.3 & IoU=0.5 \\
\hline\hline
MLVI~\cite{xu2019multilevel} & 45.30 & 27.70 & 75.70 & 59.20\\
TMLGA~\cite{ghosh2019excl} & 51.28 & 33.04 & - & -\\
CBP~\cite{wang2020temporally} & 54.30 & 35.76 & 77.63 & 65.89\\
ABLR~\cite{yuan2019find} & 55.68 & 36.79 & - & -\\
2D-TAN~\cite{2DTAN_2020_AAAI} & 56.92 & 42.08 & \underline{82.64} & 73.01\\
DRN~\cite{zeng2020dense} & - & \bf 43.95 & - & \bf 74.87\\
LGI~\cite{mun2020local} & \underline{58.52} & 41.51 & - & -\\

DeNet(ours) &\bf 61.93 & \underline{43.79} & \bf 86.02 & \underline{74.13} \\
\hline
\end{tabular}
}
\end{center}
\caption{Comparison with state-of-the-art methods on ActivityNet Captions (combination of two val\_sets) using standard metrics; bold font indicates best results, underlined second-best.}
\label{tab:2}
\end{table}

\subsection{Comparison with state-of-the-arts}
First of all, we compare our model DeNet with other state-of-the-art methods using standard metrics on two datasets, which contains CTRL~\cite{gao2017tall}, SMRL~\cite{wang2019language}, MAC~\cite{ge2019mac}, MLVI~\cite{xu2019multilevel}, CBP~\cite{wang2020temporally}, SAP~\cite{chen2019semantic}, MAN~\cite{zhang2019man}, 2D-TAN~\cite{2DTAN_2020_AAAI}, EXCL~\cite{ghosh2019excl}, TMLGA~\cite{rodriguez2020proposal}, DRN~\cite{zeng2020dense}, SCDM~\cite{yuan2019semantic}, LGI~\cite{mun2020local} and ABLR~\cite{yuan2019find}. Table~\ref{tab:1} and Table~\ref{tab:2} report the results on Charades-STA and ActivityNet Captions, respectively. For a fair comparison, all of the performances listed in Table~\ref{tab:2} are based on the combination of two validation sets on ActivityNet Captions. In the standard metrics, our method DeNet achieves competitive performances on both datasets, especially on the Charades-STA dataset. For example, DeNet obtains 3.04\% gains in "R@1,IoU=0.7" and 6.78\% gains in "R@5,IoU=0.7".

\begin{figure}[t]
\begin{center}
   \includegraphics[width=1\linewidth]{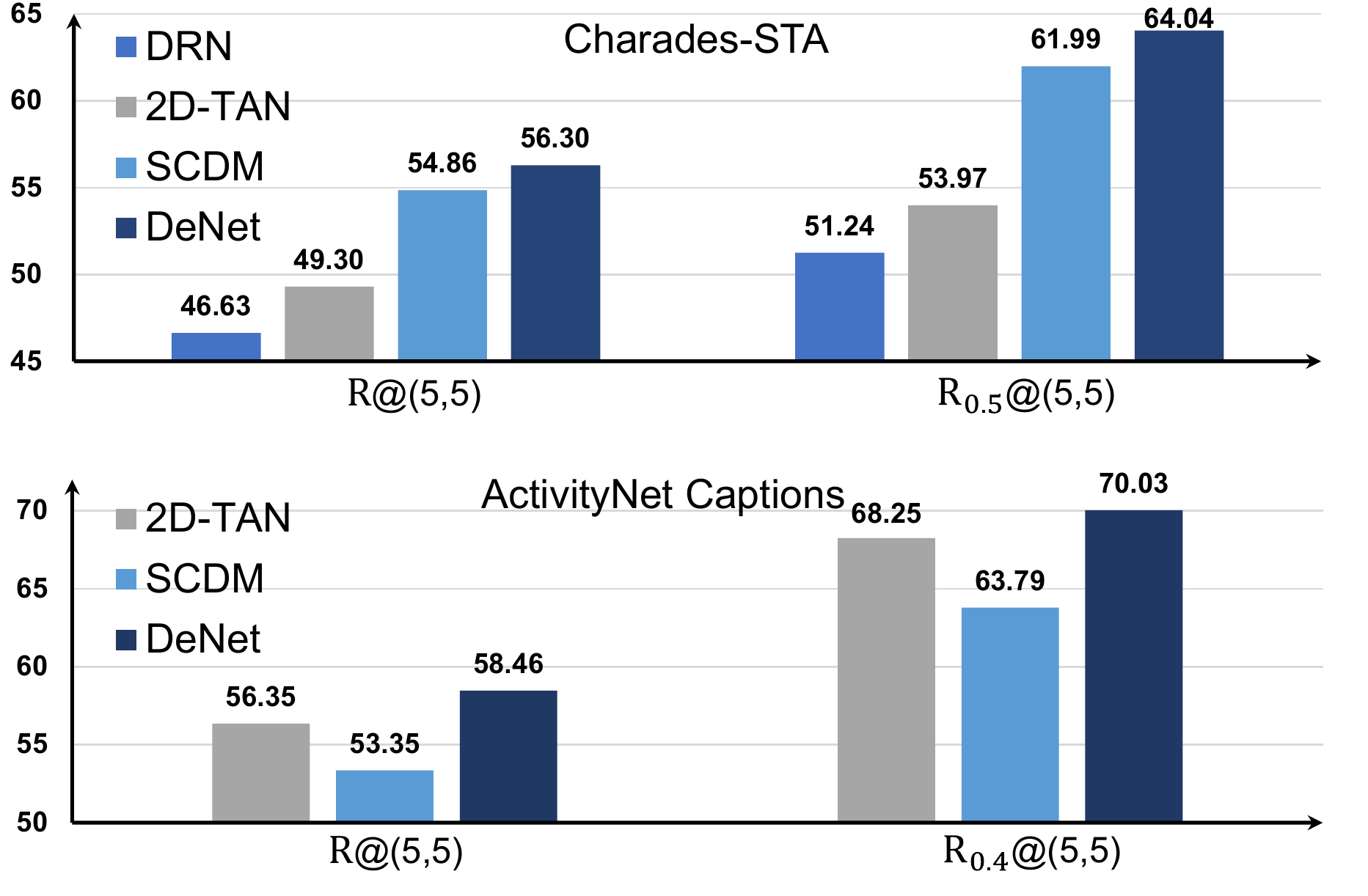}
\end{center}
   \caption{Performances on Charades-STA (top) and ActivityNet Captions (bottom) using multi-label metrics(IoU = 0.5), and at most 5 predictions and 5 annotations are taken into consideration. Best viewed in color.}
\label{fig:4}
\end{figure}

Then, to better evaluate performances of multiple predictions, we compare our model DeNet with some related methods (including 2D-TAN~\cite{2DTAN_2020_AAAI}, DRN~\cite{zeng2020dense} and SCDM~\cite{yuan2019semantic})\footnote {We test 2D-TAN and DRN using pretrained official models and SCDM using third-party implementation~\cite{otani2020uncovering}.} using "R@($N,G$), IoU=$\alpha$" and "R$_\beta$@$(N,G)$, IoU=$\alpha$". In this work, we take account of at most 5 predictions ($N$ = 5) and 5 temporal annotations ($G$ = 5). To reserve an average of 3 annotations for each query, $\beta$ is set to 0.5 on Charades-STA, and 0.4 on ActivityNet Captions. Figure~\ref{fig:4} illustrates the results. In contrast to performances in standard metrics, proposal-based methods (\ie 2D-TAN and SCDM) outperform the proposal-free method (\ie DRN) in new multi-label metrics. It means proposal-based methods tend to better cover multiple-styles annotations, yet most proposal-free models are biased to single-style annotations.  We consider it is because most proposal-free models tend to produce dense predictions. However, our proposal-free-based DeNet still outperforms the above methods on both datasets, \eg 1.75\% gains on ActivityNet Captions in terms of R@(5,5). It validates our method has an advantage in matching the multi-styles annotations.

\subsection{Ablation studies}
\noindent
{\bf Robustness for query uncertainty.} 
\begin{table}
\begin{center}
\setlength{\tabcolsep}{4pt}
{
\begin{tabular}{c|cccc}
\hline
Method & DRN~\cite{zeng2020dense} & 2D-TAN~\cite{2DTAN_2020_AAAI} & SCDM~\cite{yuan2019semantic} & DeNet \\
\hline\hline
$D_{var}$ & 0.338 & 0.365 & 0.286 & \bf 0.223 \\
\hline
\end{tabular}
}
\end{center}
\caption{Comparison of robustness for query uncertainty on Charades-STA. The lower value represents more consistent predictions for two siamese queries.}
\label{tab:3}
\end{table}
We conduct experiments to evaluate the robustness for query uncertainty. Specifically, we explored whether predictions of models can be consistent when using different queries in the same temporal moment. A subset is selected from the Charades-STA testing set, where each temporal moment contains two queries. If a moment contains more queries in the original testing set, we randomly select two queries. Finally, the subset is composed of 848 testing samples (corresponding to 1696 queries). Then, we use $D_{var} = 1- \mathrm{IoU}$ to compute the average distance between top-1 predictions of two queries. The lower value of $D_{var}$ represents more consistent predictions for the two corresponding queries. Table~\ref{tab:3} shows a comparison between DeNet with some methods. Our DeNet outperforms them by 6.3\%, which validates the robustness of our model for query uncertainty.

\noindent
{\bf Robustness for label uncertainty.} 
\begin{figure}[t]
\begin{center}
   \includegraphics[width=1\linewidth]{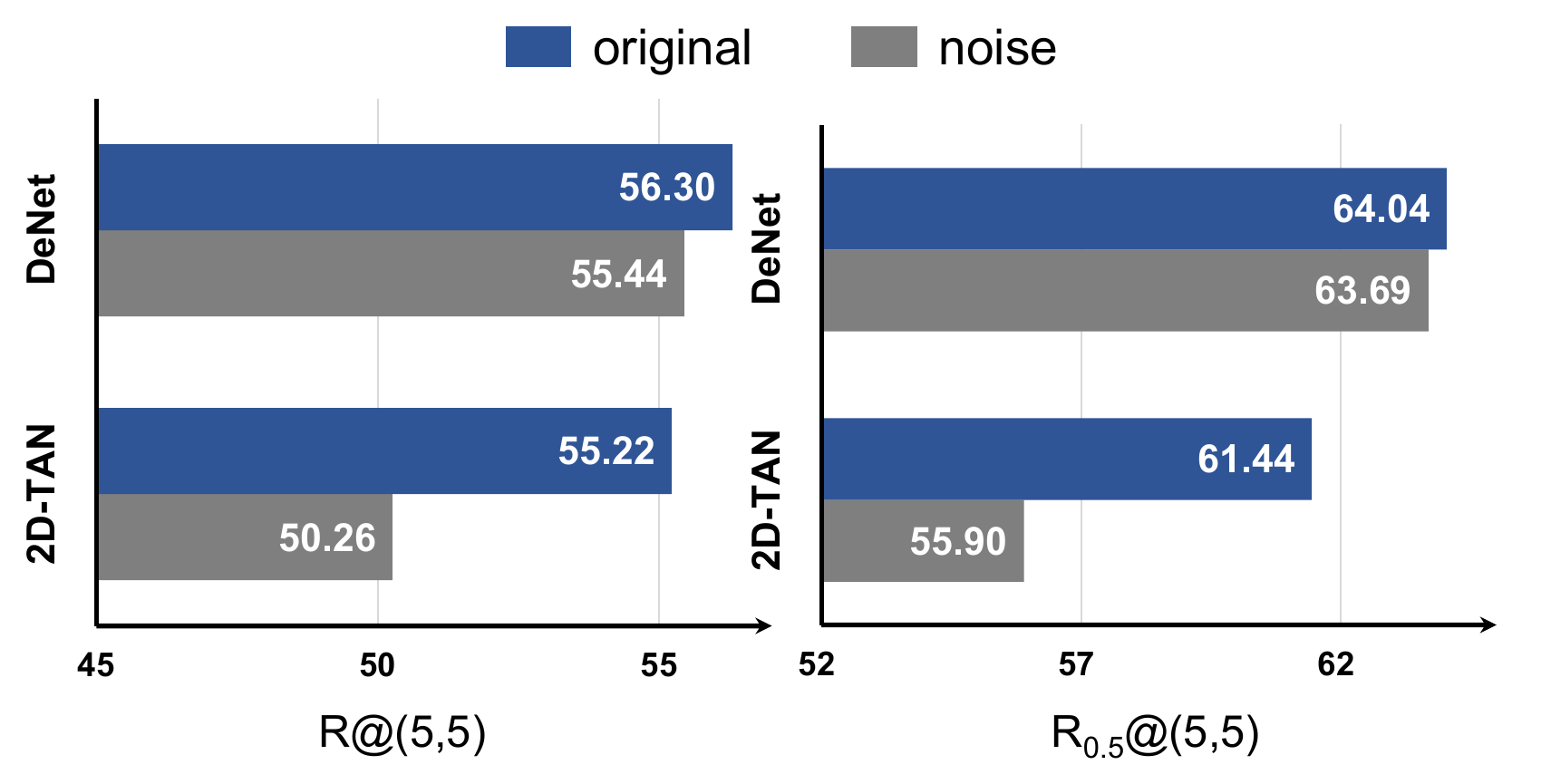}
\end{center}
   \caption{Performances of our DeNet and 2D-TAN with original annotations and noise annotations using multi-label metrics(IoU = 0.5). Best viewed in color.}
\label{fig:5}
\end{figure}
To evaluate the robustness for label uncertainty, we add perturbations in the temporal boundaries to enlarge the label uncertainty. During the training procedure, we take annotations $t_s$, $t_e$ and generate new annotations $\hat t_s = t_s + \epsilon_s(t_e-t_s)$, $\hat t_e = t_e + \epsilon_e(t_e-t_s), \epsilon_s, \epsilon_e \in \sim U(-0.5,0.5)$, where $U(-0.5,0.5)$ is uniform distribution. We train our DeNet and 2D-TAN using new annotations, then still evaluate them using original annotations. For a fair comparison, 2D-TAN adopts the same I3D feature with DeNet. Here, we mainly investigate the impact of label uncertainty on multiple predictions. Figure~\ref{fig:5} shows different results in the multi-label metrics, where "original" adopts the previous annotations, and "noise" adopts the new annotations. Compared to 2D-TAN, DeNet only drops slightly using noise annotations, \eg 0.35\% vs 5.54\% in "R$_{0.5}$@(5,5), IoU = 0.5". It also means that our method can mitigate the reliance on precise annotations in real scenarios.

\noindent
{\bf Analysis on language encoding.} 
\begin{table}
\begin{center}
\setlength{\tabcolsep}{2.6pt}
{
\begin{tabular}{l|cccc}
\hline
\multirow{2}{*}{ Method } & R@1 & R@5 & R@(5,5) & R$_{0.5}$@(5,5) \\
                         & IoU=0.5 & IoU=0.5 & IoU=0.5 & IoU=0.5 \\
\hline\hline
DeNet w/o PoS & 57.47 & 90.90 & 52.64&58.97\\
DeNet-Relation & 58.12 &\bf 91.34 & 54.32 & 61.76\\
DeNet-All & 58.23 & 88.76 & 46.20 & 50.32 \\
\hline
DeNet & \bf 59.70 &  91.24 & \bf 56.30 & \bf 64.04 \\
\hline
\end{tabular}
}
\end{center}
\caption{Ablation studies of language encoding on Charades-STA; bold font indicates best results.}
\label{tab:4}
\end{table}
In this subsection, we investigate the contribution of the language encoding under query uncertainty and set three variant implements. 1) "DeNet w/o PoS" encodes entire language without PoS. 2) "DeNet-Relation" encodes the relation feature as a Gaussian distribution rather than modified feature. 3) "DeNet-All" encodes both relation feature and modified feature as Gaussian distributions. Table~\ref{tab:4} shows the results. 

Firstly, it's more effective to disentangle language into two types of features (DeNet) than a single feature (DeNet w/o PoS). DeNet benefits from Parts-of-Speech parsing when extracting discriminative features. Secondly, for producing multiple predictions, it's more beneficial to encode the modified feature as Gaussian distribution instead of the relation feature (DeNet-Relation). Thirdly, when both two types of features are encoded as distributions (DeNet-All), it will cause performance degradation. 

\noindent
{\bf Analysis on temporal regression.} 
\begin{table}[t]
\begin{center}
\setlength{\tabcolsep}{1pt}
{
\begin{tabular}{l|cccc}
\hline
\multirow{2}{*}{ Method } & R@1 & R@5 & R@(5,5) & R$_{0.5}$@(5,5) \\
                         & IoU=0.5 & IoU=0.5 & IoU=0.5 & IoU=0.5 \\
\hline\hline
DeNet-Boundary &  57.88 & 89.25 & 55.42 & 63.14 \\
DeNet-Centerness & 57.85 & 89.17 & 54.40 & 62.18\\
\hline
DeNet-Single & 57.45 & 89.19 & 55.38 & 62.95\\
DeNet w/o min-loss & 58.90 & 69.11 & 42.18 & 50.61\\
\hline
DeNet & \bf 59.70 & \bf 91.24 & \bf 56.30 & \bf 64.04 \\
\hline
\end{tabular}
}
\end{center}
\caption{Ablation studies of temporal regression on Charades-STA; bold font indicates best results.}
\label{tab:5}
\end{table}
In this subsection, we investigate the contribution of our temporal regression under label uncertainty. Firstly, we set two variant implements to validate the benefit of predicting the center-width as an auxiliary head. 1) "DeNet-Boundary" only predicts the start-end boundary. 2) "DeNet-Centerness" only predicts the center-width. As shown in Table~\ref{tab:5}, when supervised from two perspectives, our model DeNet can obtain gains in terms of all metrics.

Secondly, we set two variant implements to investigate settings of two independent branches. 1) "DeNet-Single" represents that we only build a single-output branch. 2) "DeNet w/o min-loss" replaces $\mathcal{L}_{multi}$ with $\mathcal{L}_{single}$ for multi-output branch. Table~\ref{tab:5} summarizes different results. The original DeNet with two independent regression branches outperforms the model with only a single-output branch (DeNet-Single). For each sample, the single-output branch aims at matching the single-style annotations, yet the multi-output branch aims at matching potential multiple annotations. We consider the two different tasks may disturb each other once relied on one same branch. In terms of multiple predictions, performances will drop dramatically without min-loss (DeNet w/o min-loss), \eg 22.13\% drop in "R@5, IoU = 0.5". Thus, min-loss is necessary to learn multiple predictions for the multi-output branch.
\begin{figure}[th]
\begin{center}
% \fbox{\rule{0pt}{1.5in} \rule{0.9\linewidth}{0pt}}
   \includegraphics[width=0.95\linewidth]{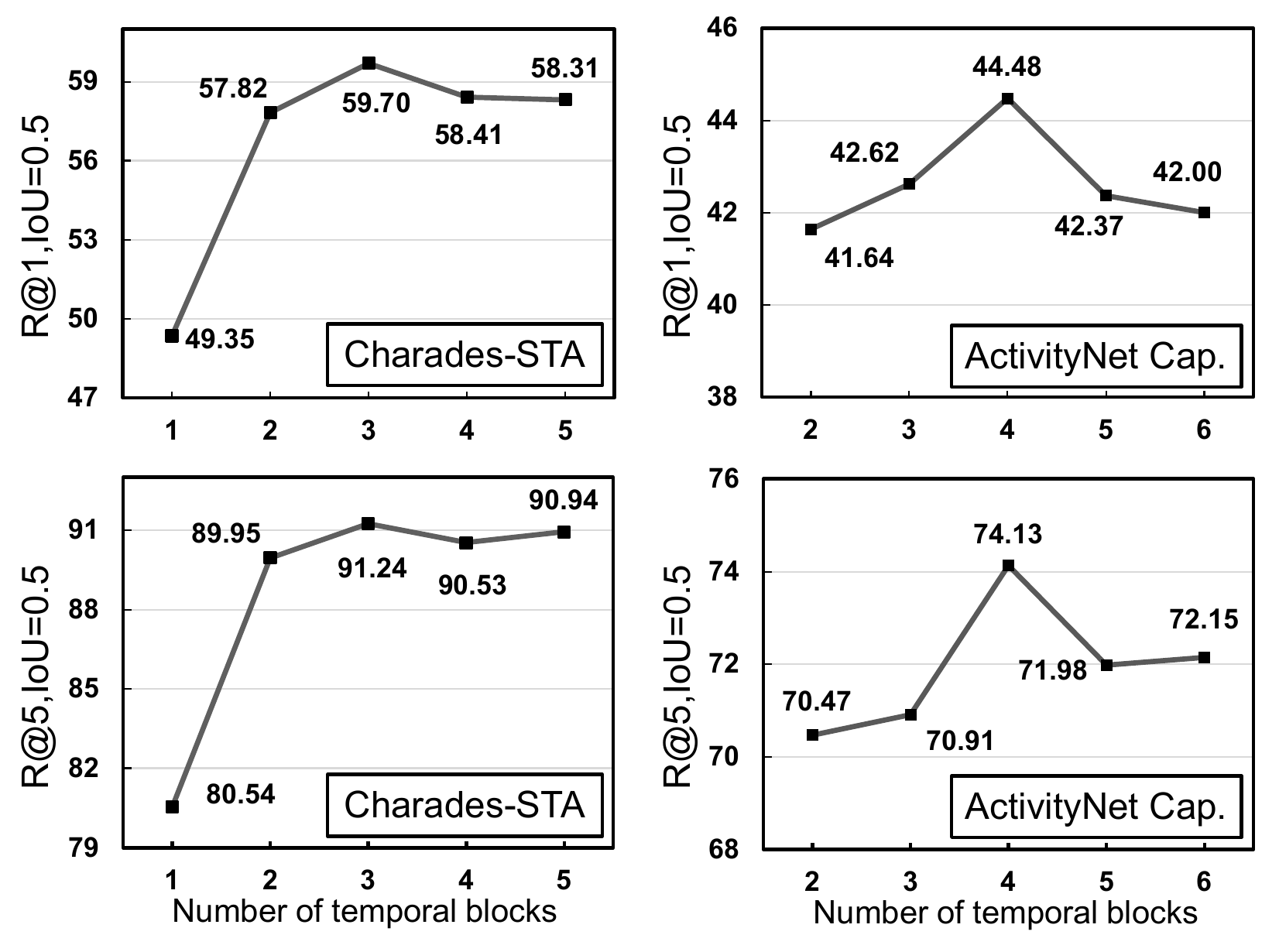}
\end{center}
   \caption{Effect of the number of stacked temporal blocks on Charades-STA and ActivityNet Captions.}
\label{fig:6}
\end{figure}

Thirdly, we analyze the impact of the number of stacked temporal blocks. Each temporal block contains a Temporal Convolutional layer and a Multi-head Attention layer. Figure~\ref{fig:6} shows results on the Charades-STA and ActivityNet Captions. We observe that our proposed method DeNet achieves best performances when the number of stacked temporal blocks reaches 3 for Charades-STA and 4 for ActivityNet Captions. 
We consider that fewer temporal blocks can not capture the long-range temporal dependencies, yet more temporal blocks may face over-fitting risk.

\begin{figure}[th]
\begin{center}
   \includegraphics[width=1.05\linewidth]{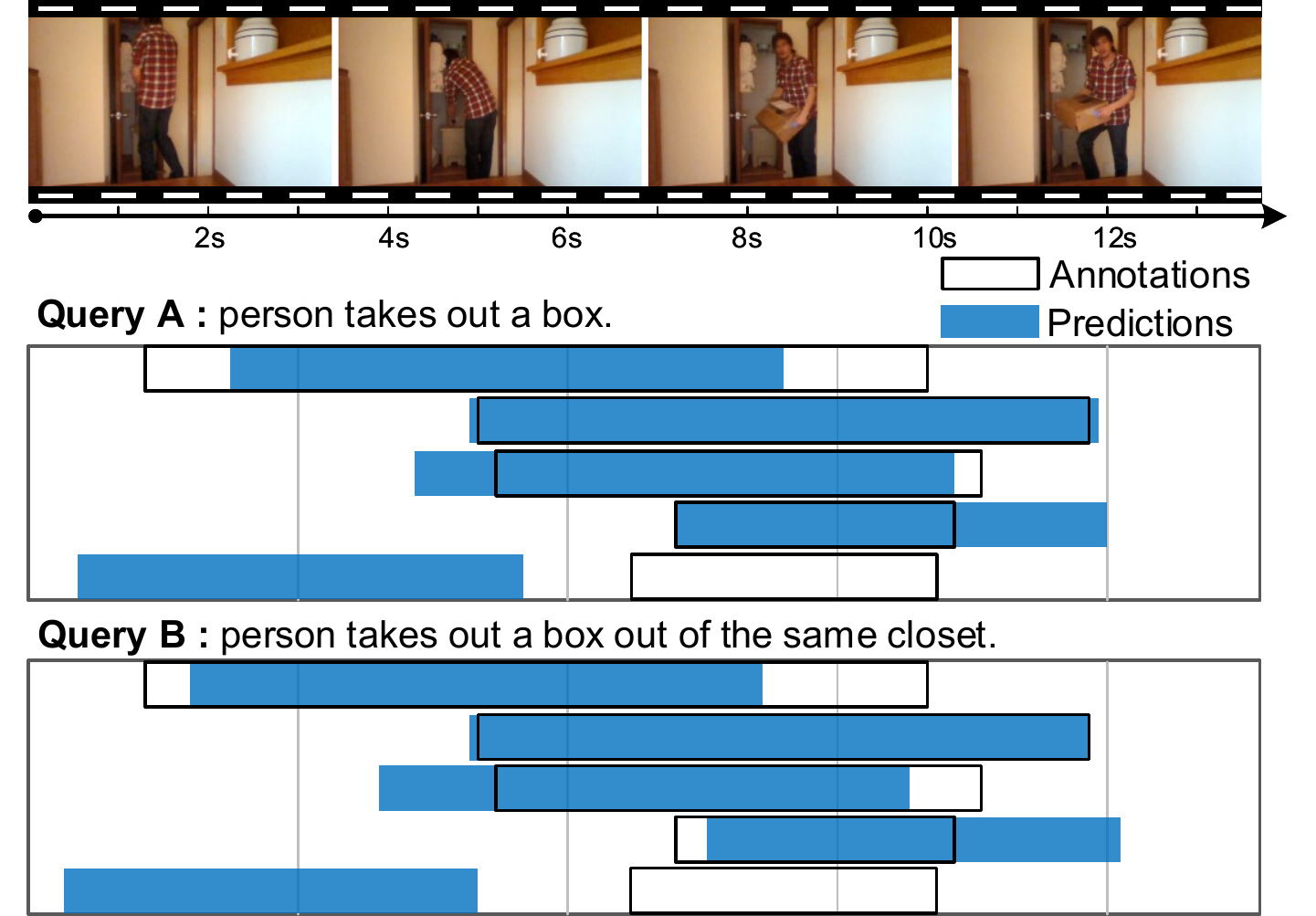}
\end{center}
   \caption{Qualitative results on Charades-STA dataset.}
\label{fig:7}
\end{figure}

\noindent
{\bf Qualitative results.} 
Figure~\ref{fig:7} illustrates multiple predictions generated by DeNet. We can find the temporal boundaries of different annotations exist disagreement for the same query. For the same query, the multiple predictions generated by DeNet can match each annotation as much as possible. For the same event, predictions of different queries (\ie Query A and Query B) tend to be consistent.

%-------------------------------------------------------------------------
\section{Conclusion}
In this paper, we propose DeNet to embrace human uncertainty for temporal grounding. Firstly, DeNet adopts a decoupling method to decompose each query into relation feature and modified feature by PoS, where consistent query information and expression variance can be obtained respectively. Then, DeNet uses a de-bias mechanism to produce diverse yet plausible predictions, aims to mitigate the reliance on single-style annotations. Experiments on two datasets validate its effectiveness and robustness.

\section{Acknowledgments}
This work was partly funded by the National Key Research and Development Program (2017YFB1002401), NSFC(No.61971281), and STCSM(18DZ2270700).

{\small
\bibliographystyle{ieee_fullname}
\bibliography{egbib}
}

\end{document}